\documentclass[11pt]{article}

\usepackage[final]{acl}
\usepackage{kotex}
\usepackage{times}
\usepackage{latexsym}
\usepackage{multirow}
\usepackage{array}
\newcolumntype{P}[1]{>{\raggedright\arraybackslash}p{#1}}
\usepackage{url} 
\usepackage{makecell} 
\usepackage{xcolor}
\usepackage{pifont}
\usepackage{stackengine}
\usepackage{tikz}
\usepackage{threeparttable}
\usetikzlibrary{calc, decorations.pathmorphing}
\newcommand{\insertsp}{%
    \tikz[overlay, remember picture, baseline=-0.3ex] {%
        \node[inner sep=0pt, font=\bfseries, text=red, yshift=0.8ex] {$\vee$};%
    }%
    \hspace{0.05em}
}
\usepackage{threeparttable}

\newcommand{\connectsp}[2]{%
    #1%
    \hspace{0.0em}
    \textcolor{red}{%
        \tikz[baseline=-1.3ex] \draw[thick, line cap=round] (0,0.1ex) arc (180:0:0.1em);%
    }%
    \hspace{0.05em}%
    #2%
}

\hypersetup{
  hidelinks
}

\usepackage[T1]{fontenc}
\usepackage[T5]{fontenc}
\usepackage[utf8]{inputenc}

\usepackage{microtype}
\usepackage{comment}

\usepackage{inconsolata}

\usepackage{graphicx}
\usepackage{amsmath}
\usepackage{tabularx}
\usepackage{booktabs} 

\title{Pushing the Boundaries of Multiple Choice Evaluation to One Hundred Options}

\author{Nahyun Lee \\
  Chung-Ang University \\
  \texttt{naa012@cau.ac.kr} \\\And
  Guijin Son \\
  Seoul National University \\
  \texttt{guijin.son@snu.ac.kr} \\}

\begin{document}
\maketitle
\begin{abstract}

Multiple choice evaluation is widely used for benchmarking large language models, yet near ceiling accuracy in low option settings can be sustained by shortcut strategies that obscure true competence.
Therefore, we propose a massive option evaluation protocol that scales the candidate set to one hundred options and sharply reduces the impact of chance performance.
We apply this framework to a Korean orthography error detection task where models must pick the single incorrect sentence from a large candidate set.
With fixed targets and repeated resampling and shuffling, we obtain stable estimates while separating content driven failures from positional artifacts.
Across experiments, results indicate that strong performance in low option settings can overstate model competence. 
This apparent advantage often weakens under dense interference at high $N$, revealing gaps that conventional benchmarks tend to obscure.
We identify two failure modes, semantic confusion and position bias toward early options under uncertainty.
To isolate the effect of context length, we run padding controlled and length matched tests, which suggest that the main bottleneck is candidate ranking rather than context length.
Together, these findings support massive option evaluation as a general framework for stress testing model reliability under extreme distractor density, beyond what low option benchmarks can reveal.

\end{abstract}

\section{Introduction}

Multiple Choice Question Answering (MCQA) has emerged as the dominant evaluation format for benchmarking Large Language Models (LLMs), offering a scalable and cost-effective means to assess everything from mathematical reasoning to multilingual nuance~\citep{hendrycks2020measuring, lai2017race, ling2017program, talmor2019commonsenseqa, zellers2019hellaswag}.
However, as models approach the performance ceiling on widely used benchmarks, a fundamental structural limitation has become increasingly apparent~\citep{achiam2023gpt, bowman2024eight}.
In traditional settings where the number of options is small ($N \in \{4, 5\}$), high accuracy often reflects a deceptive mixture of genuine knowledge and short-cut strategies~\citep{robinson2023larp, gudibande2023false, liang2022holistic, chang2024survey}.
Within such narrow search spaces, models can flourish by using partial elimination, distributional guessing, or by exploiting incidental formatting cues, which make it nearly impossible to discern if a model truly \emph{knows} the answer or is simply navigating a limited field of distractors~\citep{wang2024large,pezeshkpour2023large}.

Although open-ended evaluations can avoid some pitfalls of low-option MCQA, they impose substantial scoring burdens. As a scalable alternative, we introduce a massive-option environment with $N=100$~\citep{zheng2023judging}, which effectively turns evaluation into a high-precision selection problem under dense candidate interference. 
Scaling to one hundred options sharply reduces chance performance from 25\% to 1\%, making near-ceiling results at low $N$ less informative and exposing competence that is otherwise obscured by narrow search spaces.

We demonstrate our approach on a Korean orthography error identification task. By requiring models to distinguish a single error from highly similar correct candidates, this task serves as a rigorous test of fine-grained linguistic knowledge.

Our investigation finds a startling divergence in model robustness masked by conventional benchmarks. 
While frontier systems (\textsc{Gemini} family) maintain high reliability, other models like \textsc{HyperCLOVAX-Think} and \textsc{EXAONE-4.0}, which exhibit near-perfect performance at low option counts, suffer catastrophic degradation at $N=100$, with accuracy dropping by roughly $70$ percentage points in the most challenging environments.
Crucially, this collapse is not merely a failure of retrieval but often marks a regression to primitive decision heuristics where models abandon content analysis in favor of positional shortcuts such as selecting the first few options~\citep{liu2024lost,turpin2023language,pezeshkpour2023large}.
To isolate the mechanisms of model failure, we maintain a fixed set of target errors while systematically varying the difficulty of the distractor pool and employing repeated permutations of option order.

Our study offers three primary contributions to the field of LLM evaluation.
First, we demonstrate that scaling to one hundred options produces substantially stronger model differentiation, revealing performance gaps that remain invisible near the ceiling of conventional benchmarks.
Second, through padding-based controls, we prove that the observed performance collapse at high $N$ is driven by the complexity of candidate ranking rather than token length or long-context limitations alone. 
Third, we introduce a suite of quantitative tools for deep analysis, including the Bubble Index to measure low-option competence inflation, and positional diagnostics to quantify when models fall back to early-option selection under pressure.

Ultimately, this work provides a practical methodology for evaluating \emph{true competence}, the ability to deterministically identify the correct option amid heavy interference. 
More broadly, the results suggest that next-generation benchmarks should complement broad subject coverage with stress tests that probe reliability under dense, confusable alternatives, since this regime exposes failure modes that low-option settings often conceal.

\section{Related Works}
\subsection{Evolution of MCQA: From Low-N to Massive-Option}Multiple-Choice Question Answering (MCQA) is the standard for evaluating Large Language Models (LLMs) due to its scalability \citep{hendrycks2020measuring, huang2023c}. In the Korean domain, benchmarks like HAE-RAE Bench~\citep{son2024hae}, KMMLU~\citep{son2025kmmlu}, KoBALT~\citep{shin2025kobalt} and KoBEST~\citep{kim2022kobest} have adopted the $4$- or $5$-option format to assess linguistic and cultural competence.
However, recent studies warn that such low-$N$ settings allow models to exploit probabilistic heuristics rather than demonstrating genuine reasoning \citep{zheng2024large, wang2023robustness}.
By scaling to $N=100$, our work neutralizes these elimination-based shortcuts and forces a shift toward a more deterministic retrieval regime, addressing the latent fragility of standard benchmarks.

\subsection{Robustness Against Information Interference}
The impact of distractors is a core challenge in LLM robustness. 
Even minimal irrelevant context can significantly degrade reasoning performance \citep{jia2017adversarial, shi2023large}, while semantically relevant distractors can cause performance drops exceeding 45\% \citep{wang2025adaptive}. 
While existing research primarily focuses on small-scale adversarial injections, our massive-option environment creates a cognitive stress test. 
This setup simulates a needle-in-a-haystack scenario~\citep{kamradt2023needle} within the context window, testing the model's ability to suppress interference from a high density of competing candidates.

\subsection{Positional Bias and Decision Complexity}
LLMs exhibit well-documented positional biases, such as the \textit{Lost in the Middle} phenomenon, where retrieval accuracy dips for information in the center of a sequence \citep{liu2024lost}. 
In MCQA, this manifests as \textit{Selection Bias}, where models favor specific option IDs regardless of content \citep{pezeshkpour2023large, robinson2023larp}.
While calibration methods address these in low-$N$ settings~\citep{zhao2021calibrate}, their efficacy in massive-option regimes is underexplored. 
Our study employs padding control experiments to decouple context length from option count, revealing that performance collapse at $N=100$ is driven by \textit{Positional Fallback} strategies under high decision complexity, rather than simple token length limitations.

\section{Experiments}

\begin{table*}[t]
\centering
\small
\renewcommand{\arraystretch}{1.25}
\setlength{\tabcolsep}{8pt}

\begin{tabular}{p{2cm} p{4cm} p{3cm} p{3cm} c}
\toprule
\textbf{Category} & \textbf{Sub-category} & \textbf{Incorrect (Examples)} & \textbf{Correct (Examples)} & \textbf{Count} \\
\midrule

\multirow{4}{*}{\textbf{Spacing Errors}}
  & Particles (Auxiliary/Case)
  & \connectsp{기차}{보다}, \connectsp{하늘}{만큼}
  & 기차보다, 하늘만큼
  & 4 \\
  & Dependent Nouns
  & 시간\insertsp만에, 들었을\insertsp뿐
  & 시간 만에, 들었을 뿐
  & 4 \\
  & Predicates \& Endings
  & \connectsp{떨어질}{망정}
  & 떨어질망정
  & 3 \\
  & Lexicalized Compounds
  & \connectsp{못지}{않게}, 수\insertsp밖에
  & 못지않게, 수밖에
  & 4 \\
\midrule

\multirow{3}{*}{\shortstack[l]{\textbf{Orthographic}\\\textbf{Errors}}}
  & Sai-siot \& Compounds
  & \textcolor{red}{뒷}처리, \textcolor{red}{싯}가
  & 뒤처리, 시가
  & 4 \\
  & Phonological Confusions \newline (Vowels, Morphologies)
  & \textcolor{red}{왠}일, 금\textcolor{red}{새}, 희\textcolor{red}{안}한
  & 웬일, 금세, 희한한
  & 6 \\
  & Standard Word Form Error
  & 일일\textcolor{red}{히}, 함부\textcolor{red}{러}
  & 일일이, 함부로
  & 5 \\
\bottomrule
\multicolumn{4}{r}{\textbf{Total}} & \textbf{30} \\
\end{tabular}

\caption{\textbf{Distribution and examples of error types} in the 30 target sentences used for experimentation. 
The dataset covers representative spacing and orthographic error patterns in Korean.}
\label{tab:target-sentence-type}
\end{table*}

\subsection{Task Formulation}
We adopt an \textit{Error Identification} task where the model must select the single orthographically incorrect sentence from a pool of otherwise correct candidates. 
We prioritized this specific domain because it presents a demanding test of fine-grained linguistic discrimination. 
Unlike semantic tasks that often entail subjective interpretation, orthography offers objective verifiability while ensuring high surface similarity between the target and distractors. 
This structural characteristic forces the model to abandon superficial pattern matching in favor of exact knowledge retrieval. 
While our immediate focus remains on orthography, this massive-option discriminator framework is inherently extensible to other high-precision domains including syntax debugging in code generation and factual consistency checking in knowledge retrieval.

\subsection{Dataset Construction}
\subsubsection*{Dataset Pool}
The dataset pool comprises 750 Korean sentences drawn from the National Institute of Korean Language (NIKL).
To reduce lexical and stylistic mismatch between targets and distractors, we retrieve base sentences using approximately 100 high-confusion keywords associated with frequent orthographic error patterns (e.g., dependent nouns and auxiliary predicates), and generate perturbed variants from these bases.

\subsubsection*{Validation and Labeling}
Orthographic correctness is operationalized via an ensemble of three Korean spell checkers: 
\textsc{Daum}\footnote{\url{https://dic.daum.net/grammar_checker.do}},
\textsc{Saramin}\footnote{\url{https://www.saramin.co.kr/zf_user/tools/character-counter}},
and the \textsc{Nara}\footnote{\url{https://nara-speller.co.kr/speller/}}.
A strict unanimity criterion is used for labeling.
A sentence is retained as a \textit{distractor} only if none of the three tools flags it as erroneous.
A perturbed sentence is retained as a \textit{target} only if all three tools unanimously flag it as erroneous.

Items without unanimity are discarded to avoid borderline cases and to improve label precision.
We filter out cases governed by exception-heavy conventions, especially convention-sensitive spacing decisions and acceptable orthographic variants (e.g., cases where both spacing and concatenation are widely regarded as permissible).
This conservative protocol prioritizes precision over coverage and yields a high-confidence set of targets and distractors for controlled evaluation.
Additional details on the labeling and exclusion rules are provided in Appendix~\ref{sec:appendix_ensemble}.

\subsubsection*{Difficulty Scoring and Distractor Environments}
We assign each sentence a four-level difficulty score via a deterministic heuristic over morpho-syntactic surface features (e.g., eojeol/morpheme length, clause-ending counts, high-confusion indicators) and discretize by quantiles (Levels 1--4).
We define two distractor environments that differ only in the distractor pool:
\textit{Easy} samples from Levels 1--2, and \textit{Full} samples from Levels 1--4.
See Appendix~\ref{sec:appendix_difficulty} for scoring details and examples.


Using this score, we define two distractor environments that differ only in the distractor pool:
the \textit{Easy} environment samples distractors from Levels 1--2, and the \textit{Full} environment samples distractors from Levels 1--4.

\subsubsection*{Target Generation}
We select 30 representative base sentences from the \textit{Full} environment and generate erroneous targets via controlled perturbations reflecting common spacing and orthographic error patterns (Table~\ref{tab:target-sentence-type}).
All targets satisfy the unanimity criterion.
Consequently, each $N$-option instance contains one verified error (the \textit{target}) and $N-1$ verified-correct distractors.


\subsection{Models}
We evaluate frontier multilingual and Korean-specialized LLMs:
\textsc{Gemini} (\textsc{2.5-Flash}, \textsc{3-Pro-Preview})~\citep{comanici2025gemini},
\textsc{A.X-4.0}~\citep{SKTAdotX4},
\textsc{EXAONE-4.0-32B}~\citep{exaone-4.0},
and \textsc{HyperCLOVAX-SEED-Think-14B}~\citep{team2025hyperclova}.
All evaluations are zero-shot with temperature $0$; models output a single integer index, scored by exact match.

\subsection{Evaluation Protocol}
\label{sec_eval_protocol}
Each model is evaluated in a massive-option MCQA setting with option sizes $N \in \{4, 10, 20, 50, 100\}$.
To strictly isolate the impact of distractor interference, we maintain a constant set of $30$ target items while systematically varying the accompanying distractor pool.
For every target and option size $N$, we conduct $K=1{,}000$ independent evaluation trials.
In each trial, we dynamically sample $N-1$ distractors from the environment-specific pool (\textit{Easy} or \textit{Full}) and randomly shuffle the placement of the target option among them.
To ensure comparability across targets, we employed a synchronized seeding strategy where the $k$-th trial for every target utilized the same random state for distractor sampling and position shuffling.
This design ensures that all targets are subjected to statistically equivalent interference patterns.

We report the aggregated accuracy across these $1{,}000$ trials, effectively marginalizing distractor composition effects and positional artifacts to yield a robust Monte Carlo estimate of competence.

\subsection{Uncertainty Estimation}
\label{sec_uncertainty}
Uncertainty is quantified from two sources, separating sampling variability from target heterogeneity.
First, we estimate \textit{sampling variability} by constructing 95\% confidence intervals over the $K=1{,}000$ independent trials for each target.
This interval captures the model's sensitivity to the specific composition of distractors and the position of the answer within the option list.
Second, we estimate \textit{target heterogeneity} by averaging performance across the $1{,}000$ trials for each target to obtain a stable target-level score, and then computing confidence intervals across the distinct target sentences.
This dual approach clarifies whether performance fluctuations are driven by the stochasticity of distractor interference or by the intrinsic complexity of the target instances.

\section{Methods}
\label{sec_methods}

We introduce two diagnostics that track how model behavior changes as the option count increases. 
The first measures competence inflation when moving from low-option to massive-option evaluation. 
The second quantifies positional fallback under uncertainty, capturing the tendency to default to early options.

\paragraph{Notation and Accuracy.}
Let $t \in T$ represent the target items with $|T| = 30$, and let $k \in \{1, \ldots, K\}$ denote the independent evaluation trials for each target, with $K = 1,000$ unless otherwise specified. For each trial, we record a binary correctness indicator $c^{(N)}_{t,k} \in \{0, 1\}$, the gold position $g^{(N)}_{t,k} \in \{1, \ldots, N\}$, and the model response position $r^{(N)}_{t,k} \in \{1, \ldots, N\}$. 
We obtain per-target accuracy by aggregating over the independent trials
\begin{equation}
A^{(N)}_{t} = \frac{1}{K} \sum_{k=1}^{K} c^{(N)}_{t,k}.
\end{equation}

and compute the overall accuracy as the mean across all targets
\begin{equation}
A^{(N)} = \frac{1}{|T|} \sum_{t \in T} A^{(N)}_{t}.
\end{equation}

\paragraph{Chance-Normalized Competence and Bubble Index.}
Raw accuracy is not directly comparable across option counts because the random-guess baseline scales as $1/N$. We therefore report \textit{chance-normalized accuracy}
\begin{equation}
NA^{(N)} = \frac{A^{(N)} - \frac{1}{N}}{1 - \frac{1}{N}},
\end{equation}
which rescales performance so that $0$ corresponds to random guessing and $1$ to perfect accuracy.

To quantify the extent to which low-option performance inflates relative to massive-option performance, we define the \textit{Bubble Index} as
\begin{equation}
BI = 1 - \frac{NA^{(100)}}{NA^{(4)} + \epsilon},
\end{equation}
where $\epsilon$ is a small constant for numerical stability. We set $\epsilon = 10^{-6}$ in all experiments.
$BI$ equals $0$ when $NA^{(100)}$ matches $NA^{(4)}$ up to $\epsilon$, and increases as high-$N$ competence collapses relative to low-$N$.
A larger $BI$ indicates a larger gap between low-$N$ and high-$N$ chance-normalized competence, consistent with a \textit{competence bubble} that bursts under dense interference.

\paragraph{Positional Fallback under Uncertainty.}
To diagnose whether models revert to positional heuristics in high-interference regimes, we analyze the distribution of selected option indices.
We estimate the model response-position distribution at option size $N$ as
\begin{equation}
P_{\mathrm{resp}}^{(N)}(i) = \Pr\!\left[r_{t,k}^{(N)} = i\right],
\end{equation}
computed empirically over all trials $(t,k)$.
Because our dynamic sampling protocol introduces stochastic variation in target placement, the realized distribution of correct answers may deviate slightly from perfect uniformity.
We account for this by computing the \emph{empirical} gold-position distribution
\begin{equation}
P_{\mathrm{gold}}^{(N)}(i) = \Pr\!\left[g_{t,k}^{(N)} = i\right],
\end{equation}
and using it as a placement-aware baseline.
We then summarize the concentration of responses in early options using the \textit{Primacy-Fallback Index (PFI)}
\begin{equation}
PFI_{k_{\mathrm{pre}}}^{(N)} = \sum_{i=1}^{k_{\mathrm{pre}}} P_{\mathrm{resp}}^{(N)}(i),
\end{equation}
where $k_{\mathrm{pre}}$ denotes the cutoff rank defining the early portion of the option list.
For instance, we set $k_{\mathrm{pre}} = 10$ when $N=100$ to monitor whether responses are disproportionately concentrated within the first 10 positions.
To isolate model bias from placement artifacts, we report \textit{Excess Primacy} as the difference between the response and gold accumulations
\begin{equation} \begin{aligned} 
\Delta PFI_{k_{\mathrm{pre}}}^{(N)} &=\sum_{i=1}^{k_{\mathrm{pre}}}\Big(P_{\mathrm{resp}}^{(N)}(i)-P_{\mathrm{gold}}^{(N)}(i)\Big) \\[-2pt]&= PFI_{k_{\mathrm{pre}}}^{(N)} - \sum_{i=1}^{k_{\mathrm{pre}}} P_{\mathrm{gold}}^{(N)}(i). 
\end{aligned}\end{equation}
Under an ideal strategy, $\Delta PFI_{k_{\mathrm{pre}}}^{(N)}$ approaches zero, whereas substantially positive values indicate a systematic fallback toward early options.
We further investigate whether this positional preference distorts performance by calculating \textit{gold-position-conditioned accuracy}.
We partition the gold positions $g_{t,k}^{(N)}$ into $B$ bins and measure accuracy within each bin as
\begin{equation}
A_{\mathrm{bin}=b}^{(N)} = \Pr\!\left[c_{t,k}^{(N)}=1 \mid \mathrm{bin}\!\left(g_{t,k}^{(N)}\right)=b \right].
\end{equation}
The resulting trend is summarized by fitting a linear model $A_{\mathrm{bin}=b}^{(N)} \approx \alpha b + \beta$, where the slope $\alpha$ quantifies the dependency of correctness on option placement.
We estimate $\alpha$ using weighted least squares, with weights proportional to the number of trials in each bin, to robustly handle finite-sample variations in bin counts.
A negative slope indicates that the model becomes increasingly prone to error as the target option appears later in the list.

\section{Results}
\label{sec_results}

\begin{table*}[t]
\centering
\small
\setlength{\tabcolsep}{4.5pt}
\renewcommand{\arraystretch}{1.05}
\resizebox{\textwidth}{!}{%
\begin{tabular}{l l r r r r l r l r r r}
\toprule
Env & Model 
& Acc$_4$($\uparrow$) & Acc$_{100}$($\uparrow$) 
& NA$_4$($\uparrow$) & NA$_{100}$($\uparrow$) 
& NA$_{100}$ 95\% CI 
& BI($\downarrow$)
& BI($\downarrow$) 95\% CI 
& $\Delta$PFI$_{10}$ & Slope$_{100}$ \\
\midrule
\multirow{5}{*}{Easy}
& Gemini-3-Pro-preview   & \textbf{0.9998} & \textbf{0.9898} & \textbf{0.9998} & \textbf{0.9897} & [0.9882, 0.9912] & \textbf{0.0101} & [0.0075, 0.0126]  & 0.0046 & -0.0013 \\
& Gemini-2.5-Flash       & 0.9949 & 0.9812 & 0.9932 & 0.9810 & [0.9700, 0.9920] & 0.0123 & [-0.0003, 0.0248] & 0.0019 & -0.0005 \\
& A.X-4.0                & 0.9767 & 0.9513 & 0.9689 & 0.9508 & [0.9466, 0.9550] & 0.0186 & [0.0095, 0.0277] & -0.0014 &  0.0086 \\
& EXAONE-4.0-32B         & 0.9983 & 0.7144 & 0.9977 & 0.7115 & [0.6948, 0.7282] & 0.2869 & [0.2695, 0.3042] & 0.0186 & -0.0588 \\
& HyperCLOVAX-Think      & 0.9890 & 0.7998 & 0.9853 & 0.7978 & [0.7885, 0.8070] & 0.1904 & [0.1780, 0.2027] & 0.0909 & -0.0146 \\
\midrule
\multirow{5}{*}{Full}
& Gemini-3-Pro-preview   & \textbf{0.9881} & \textbf{0.8574} & \textbf{0.9842} & \textbf{0.8560} & [0.8348, 0.8752] & \textbf{0.1313} & [0.1124, 0.1501] & 0.0163 & -0.0008 \\
& Gemini-2.5-Flash       & 0.9730 & 0.7172 & 0.9640 & 0.7144 & [0.6893, 0.7394] & 0.2589 & [0.2297, 0.2880] & 0.0253 & -0.0084 \\
& A.X-4.0                & 0.8898 & 0.5659 & 0.8530 & 0.5615 & [0.5353, 0.5877] & 0.3418 & [0.3160, 0.3675] & 0.0149 &  0.0317 \\
& EXAONE-4.0-32B         & 0.8119 & 0.1401 & 0.7492 & 0.1314 & [0.1267, 0.1362] & 0.8245 & [0.8154, 0.8335] & 0.3120 & -0.0488 \\
& HyperCLOVAX-Think      & 0.9257 & 0.2122 & 0.9009 & 0.2043 & [0.1821, 0.2264] & 0.7732 & [0.7484, 0.7980] & 0.5263 & -0.0675 \\
\bottomrule
\end{tabular}%
}
\caption{\textbf{Core results on Massive-Option MCQA} under two distractor environments (\textit{Easy} and \textit{Full}).
We report accuracy (Acc), chance-normalized accuracy (NA), competence bubble index (BI), and positional diagnostics at $N=100$.
Positional behavior is summarized by excess primacy $\Delta PFI_{10}$,
computed relative to the \emph{empirical} gold-position distribution, and by the gold-position slope at $N=100$.}
\label{tab:main}
\end{table*}

\begin{figure}[t]
\includegraphics[width=\columnwidth]{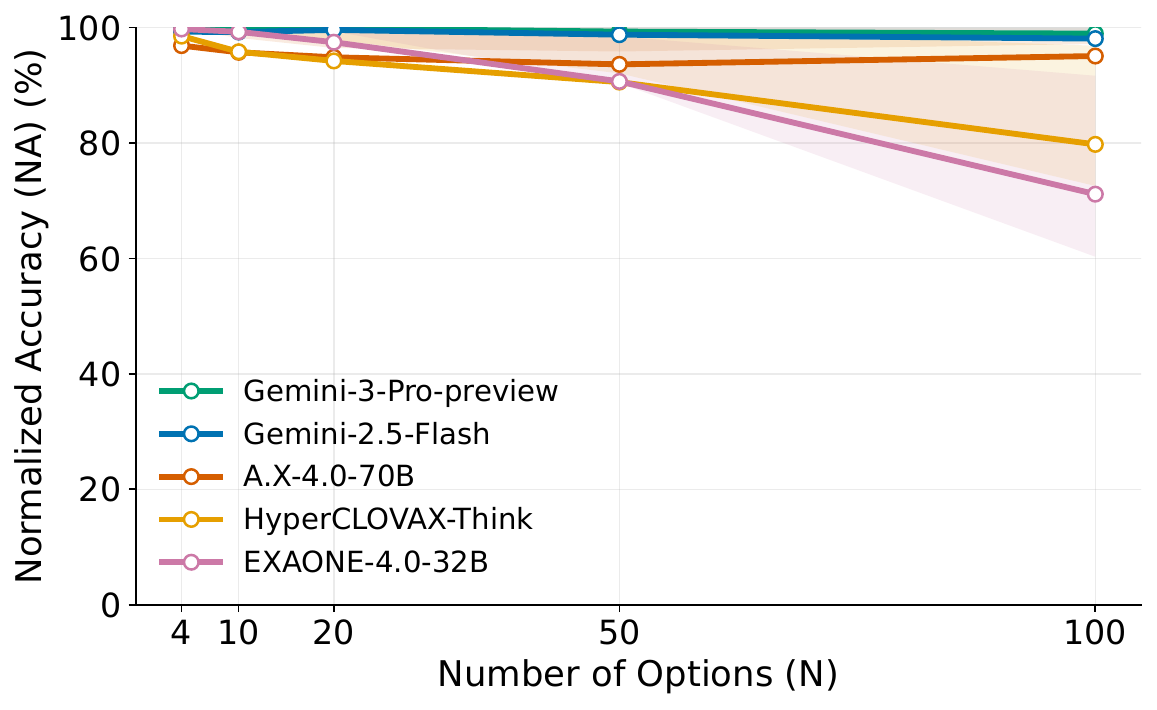}
\includegraphics[width=\columnwidth]{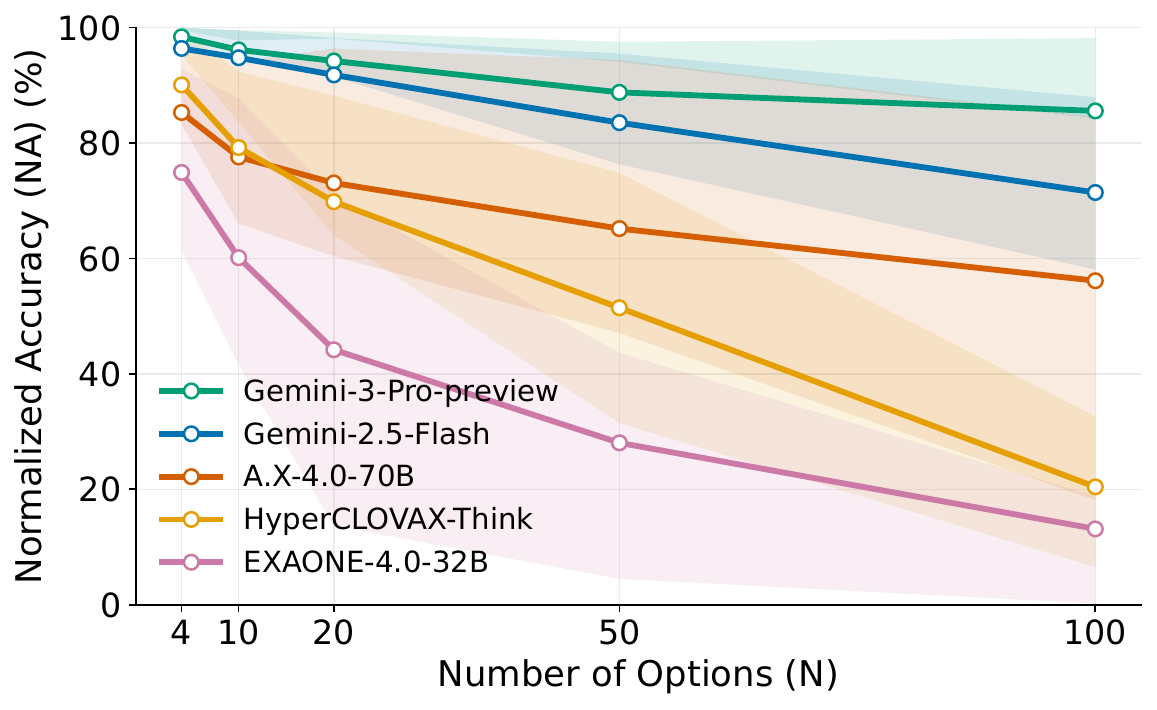}
\caption{\textbf{Chance-normalized accuracy (NA) as a function of the number of options $N$ under two distractor environments: \textit{Easy} (top) and \textit{Full} (bottom)}. 
Lines show the mean NA (\%) across 30 target sentences for each model. Shaded bands indicate the interquartile range (25--75\%) across targets, capturing variability across target sentences.}
\label{fig:overall-acc}
\end{figure}

We evaluate massive-option multiple choice under two distractor environments.
The \textit{Easy} environment draws distractors from lower difficulty levels, while the \textit{Full} environment draws distractors from the full difficulty range.
Across both environments, we keep the same set of $30$ target errors fixed and vary only distractor sampling and option placement under the dynamic protocol described in Section~\ref{sec_eval_protocol}.
Table~\ref{tab:main} reports performance at $N=4$ and $N=100$ using raw accuracy and chance-normalized accuracy (NA), together with diagnostics of competence inflation (BI) and positional behavior (excess primacy $\Delta PFI_{10}$ and gold-position slope).
Figure~\ref{fig:overall-acc} visualizes the $N$-scaling trends using $NA(N)$.
For each model and each $N$, we construct 95\% confidence intervals over the $K=1{,}000$ independent trials.
As shown in Table~\ref{tab:main}, these intervals are narrow, indicating that the aggregated performance trends are robust to the stochasticity of distractor selection.
Variation across targets is substantially larger than variation across trials, reflecting heterogeneity in target difficulty.
We therefore report the corresponding \textit{target-level} confidence intervals in Appendix Table~\ref{tab:target_level_ci}.

\subsection{Easy environment}
\label{sec_results_easy}
On the \textit{Easy} environment, models stay near ceiling at small $N$ and largely preserve their performance as $N$ increases (Figure~\ref{fig:overall-acc}, top).
Consistent with this, $BI$ values remain small for the strongest models, while some mid-tier models already show moderate competence inflation even in the Easy environment.
Positional behavior is generally limited in this regime: most models exhibit small excess primacy $\Delta$PFI$_{10}$ and near-zero gold-position slopes (Table~\ref{tab:main}).

\subsection{Full environment}
\label{sec_results_full}
The \textit{Full} environment yields a qualitatively different picture.Several models appear strong at small $N$, yet scaling to $N=100$ produces clear stratification in chance-normalized accuracy and competence inflation.
Models with small $BI$ retain most of their chance-normalized competence, whereas models with large $BI$ exhibit pronounced drops, showing that low-option accuracy can overstate robustness under dense interference.
Performance gaps at $N=100$ are accompanied by two diagnostic signatures.
In the Full environment, weaker models (e.g., EXAONE-4.0 and HyperCLOVAX-THINK) show slightly higher variability, reflected in modestly wider confidence intervals under dense interference (Table~\ref{tab:main}).
Second, these models show stronger positional confounding: they concentrate responses in early indices (larger excess primacy $\Delta$PFI$_{10}$) and exhibit more negative gold-position slopes, implying that early-position preference is coupled with gold-position-dependent correctness.
In contrast, more robust models retain higher $NA_{100}$ while remaining less sensitive to sampling variability and less confounded by option order.
Overall, massive-option evaluation separates models that primarily fail due to dense semantic interference from those that additionally exhibit positional fallback under uncertainty.
These differences help explain why models with similar low-$N$ accuracy can behave very differently at $N=100$ on the \textit{Full} environment.

\section{Long-Context Control}

\begin{figure*}[h]
    \includegraphics[width=\textwidth]{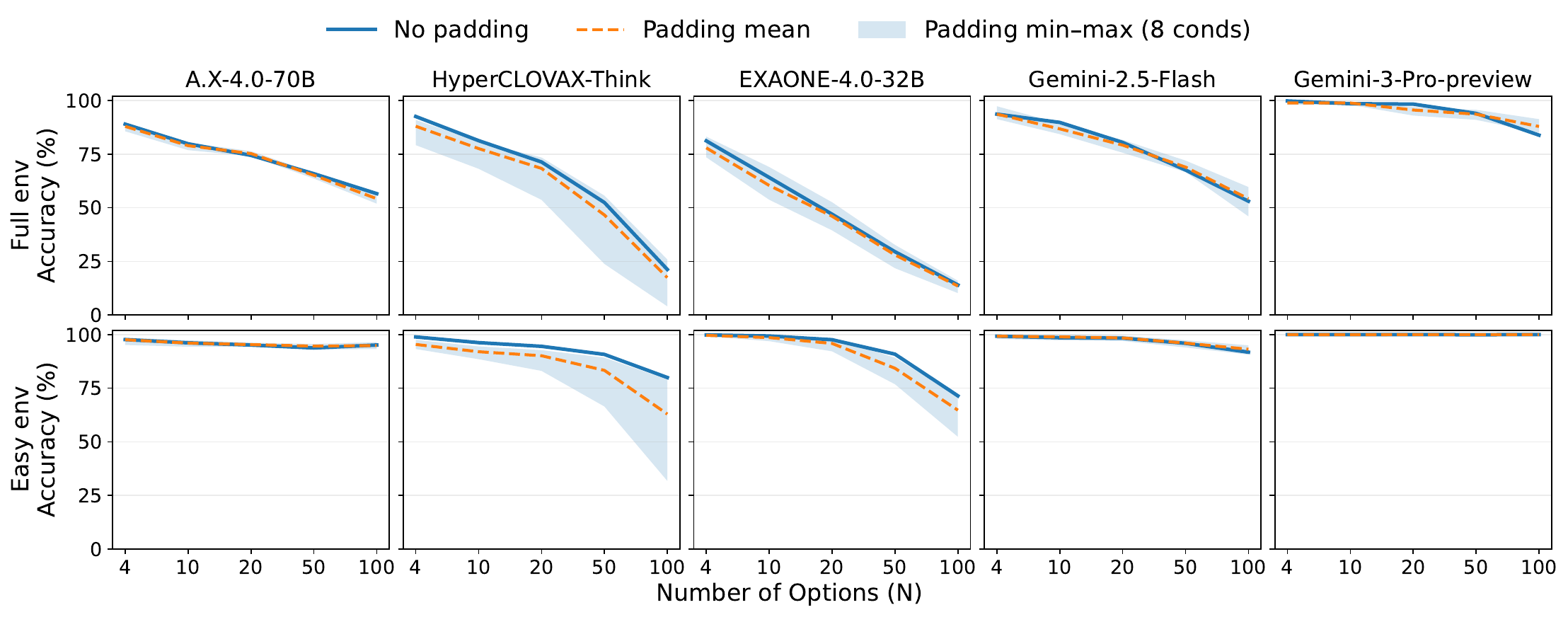}
\caption{\textbf{Padding robustness envelopes.}
No padding (solid), padding mean (dashed), and min--max over eight padding conditions (shaded) for each model and environment across \(N\in\{4,10,20,50,100\}\).}
\label{fig:padding_envelope}
\end{figure*}

\begin{table}[t]
\centering
\small
\setlength{\tabcolsep}{6pt}
\renewcommand{\arraystretch}{1.08}
\begin{tabular}{lrr}
\toprule
Model & Easy & Full \\
\midrule
Gemini-3-Pro-preview & 0.000 & 0.063 \\
Gemini-2.5-Flash     & 0.043 & 0.137 \\
A.X-4.0              & 0.032 & 0.049 \\
HyperCLOVAX-Think    & \textbf{0.473} & \textbf{0.219} \\
EXAONE-4.0           & 0.189 & 0.058 \\
\bottomrule
\end{tabular}
\caption{\textbf{Spread across paddings at $N\!=\!100$ (range).}
Each entry is $(\max\Delta-\min\Delta)$ across padding conditions, where $\Delta=\mathrm{Acc}_{100}^{\text{pad}}-\mathrm{Acc}_{100}^{\text{no-pad}}$.
Large ranges indicate strong sensitivity to padding content/placement.}
\label{tab:padding_robustness}
\end{table}

\paragraph{Padding-based length control}

A natural concern is that the drop at $N=100$ may reflect long-context limitations rather than ranking difficulty.
To isolate these factors, we conduct a padding-based control experiment that disentangles prompt-length effects from the underlying ranking challenge.
We inject task-irrelevant padding that is semantically non-informative for orthography while preserving the original question and candidate set.
The setup includes four padding types, domain-irrelevant Korean prose, English translations, symbolic noise, and enumerated lists.
By applying these paddings in both front and back positions, we obtain eight distinct conditions alongside a no-padding baseline.
As summarized in Table~\ref{tab:padding-token-counts}, this manipulation expands the effective context length to 2,000 to 5,000 tokens, comparable to or exceeding that of large-$N$ prompts.
Figure~\ref{fig:padding_envelope} and Table~\ref{tab:padding_robustness} report robustness, where each cell denotes the range of performance shifts across padding conditions, defined as $\max\Delta - \min\Delta$ at $N=100$.
Smaller ranges indicate consistent behavior regardless of padding, whereas larger ranges reflect higher sensitivity to contextual noise.
Under this measure, the \textsc{Gemini} series and \textsc{A.X-4.0} show strong stability, with spreads below $0.14$, suggesting that prompt length is a secondary factor behind the observed degradation.
Instead, the dominant difficulty stems from interference when discriminating among a dense set of confusable candidates.
A model-specific exception arises for \textsc{HyperCLOVAX-Think} and, to a lesser extent, \textsc{EXAONE-4.0} in the \textit{Easy} environment.
These models show substantial sensitivity to padding content, with particularly large degradation under English-text padding at $N=100$.
This pattern indicates that while context length is not a universal bottleneck, some models are vulnerable to specific distribution shifts in the input context.
Detailed diagnostics for \textsc{HyperCLOVAX} are provided in Appendix~\ref{sec:appendix_clovax_entext}.

\section{Decision policy under uncertainty at high $N$}
\label{sec:position}

\begin{figure}[t]
  \centering
  \includegraphics[width=\columnwidth]{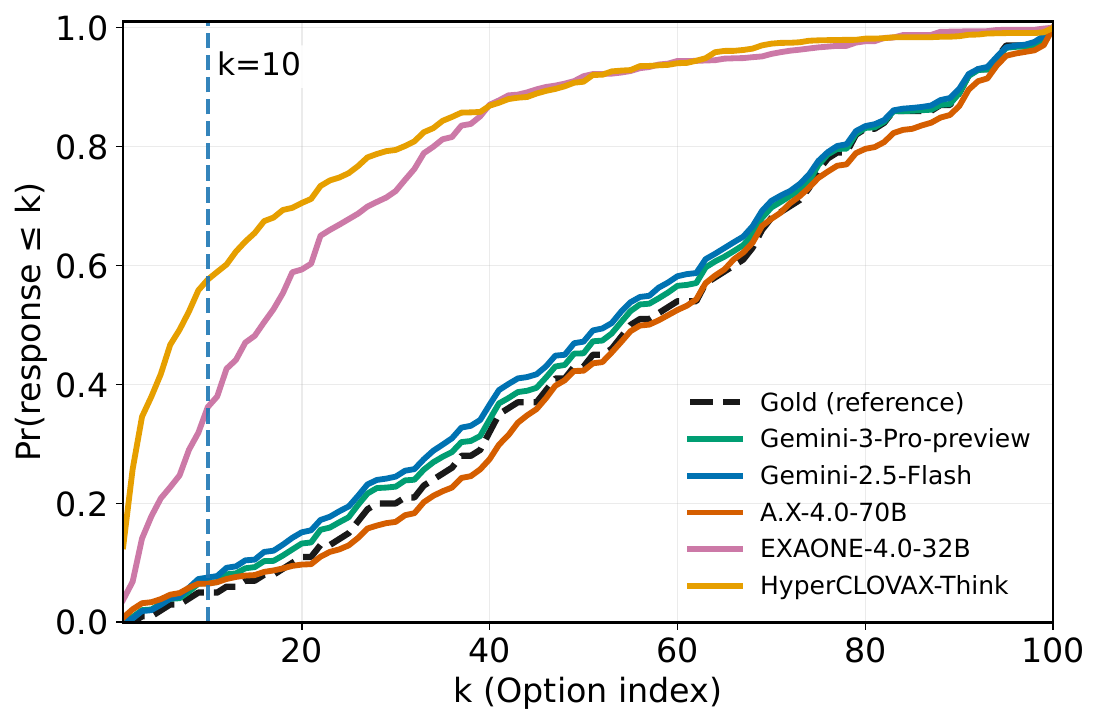}
    \caption{\textbf{Response-position CDFs at $N=100$ on the \textit{Full} environment.} The plot displays the cumulative distribution of model responses. At $k_{\mathrm{pre}}=10$, the CDF value corresponds to $PFI_{10}^{(100)}=\Pr(r\le 10)$.}
  \label{fig:resp_cdf_full}
\end{figure}

\begin{table}[t]
\centering
\small
\renewcommand{\arraystretch}{1.10}
\resizebox{\columnwidth}{!}{%
\begin{tabular}{l r r r r}
\toprule
Model & $\Delta PFI_{10}$ & Entropy$_N$ & Front$_{20}$ & Tail$_{20}$ \\
\midrule
Gemini-3-Pro-preview & 0.016 & 0.91 & 0.14 & 0.16 \\
Gemini-2.5-Flash     & 0.025 & 0.93 & 0.16 & 0.16 \\
A.X-4.0          & 0.015 & 0.92 & 0.08 & \textbf{0.23} \\
EXAONE-4.0-32B       & 0.312 & 0.79 & 0.64 & 0.01 \\
HyperCLOVAX-Think    & \textbf{0.526} & 0.73 & \textbf{0.71} & 0.02 \\
\bottomrule
\end{tabular}%
}
\caption{\textbf{Decision policy summary at $N=100$ on the \textit{Full} environment.} $\Delta PFI_{10}$ denotes excess primacy relative to the empirical gold distribution. Entropy$_N$ is the normalized entropy of the response-position distribution. Front$_{20}$ and Tail$_{20}$ represent the proportion of response mass allocated to the first and last 20 options, respectively.}
\label{tab:policy_summary}
\end{table}

We analyze decision policies in the high-$N$ regime, where dense distractors reduce candidate separability and induce uncertainty. In this setting, the central question extends beyond simple accuracy degradation to how a model allocates its choices across the option list when it lacks a clearly preferred candidate. We therefore interpret response positions as a window into the model's decision policy under uncertainty.

For each model at $N=100$, we run $K=1{,}000$ trials for each of the $|T|=30$ targets, yielding $|T|K=30{,}000$ responses in total, and aggregate the selected indices $r \in {1,\ldots,100}$.
While the expected gold position is uniform under random shuffling, the realized gold positions may deviate slightly due to finite sampling stochasticity. 
Accordingly, we utilize the empirical distribution of realized gold indices $g$ as the reference baseline and report both model and gold CDFs in Figure~\ref{fig:resp_cdf_full}. 
This CDF representation provides a stable summary of early probability mass and facilitates threshold-based diagnostics.

Figure~\ref{fig:resp_cdf_full} illustrates that models adopt qualitatively distinct decision policies in the \textit{Full} environment. 
Table~\ref{tab:policy_summary} summarizes these trends using distribution-level diagnostics. 
These include excess primacy $\Delta PFI_{10}^{(N=100)}$ which measures the surplus probability mass assigned to the first ten options, and normalized entropy $Entropy_N$ where larger values indicate dispersed responses. 
We additionally report a broader set of positional diagnostics in Appendix Table~\ref{tab:positional_appendix}, including front mean response position, the KS distance, and the gold-position slope.

Under these diagnostics, the \textsc{Gemini} models exhibit a response distribution that closely tracks the empirical gold reference. 
They maintain small $\Delta PFI_{10}^{(N=100)}$, high $Entropy_N$, and low KS distance. 
These metrics indicate that even as accuracy degrades, the response distribution does not collapse into early indices. 
Instead, choices remain broadly distributed, a behavior consistent with content-driven uncertainty rather than a position-driven fallback strategy.

Conversely, \textsc{HyperCLOVAX-Think} and \textsc{EXAONE-4.0} display a distinct policy characterized by substantial excess primacy and significantly lower entropy. 
A majority of their responses concentrate within the first $20$ options, creating large deviations from the gold reference. 
In this regime, accuracy is confounded by order because the response policy itself allocates disproportionate mass to early indices. 
\textsc{A.X-4.0} presents a divergent pattern where excess primacy remains small, yet the mean response position shifts toward the middle and tail.
This suggests a mid-to-tail preference rather than the primacy-heavy fallback observed in other models.

To examine the relationship between decision policy and difficulty directly, we present a case study in Appendix~\ref{sec:additional-pos-analysis} where evaluation trials are stratified into high-difficulty and low-difficulty groups based on accuracy. 
Figure~\ref{fig:policy_case_study} demonstrates that Gemini models preserve a high-entropy, near position-invariant policy across both groups. 
In contrast, \textsc{HyperCLOVAX-Think} and \textsc{EXAONE-4.0} retain a low-entropy, prefix-concentrated policy regardless of difficulty, indicating that their positional collapse is a persistent response strategy in the dense distractor regime.
Finally, we analyze the coupling between positional preference and correctness. As reported in Appendix Table~\ref{tab:positional_appendix}, models with strong excess primacy also exhibit more negative gold-position slopes. This finding suggests that their response policy amplifies order effects into correctness errors. The \textsc{Gemini} models, by comparison, show near-zero slopes, consistent with errors driven by candidate confusability rather than systematic order confounding.

\section{Conclusion}

This work questions the field’s continued reliance on low-option multiple choice benchmarks by introducing a massive-option evaluation protocol that probes reliability under dense interference.
Across controlled experiments, near-ceiling performance at small option counts frequently degrades at $N=100$, indicating that low-option accuracy can be maintained by shortcut strategies that do not translate to robust decision making.
Length-controlled padding tests further show that this collapse is driven primarily by the difficulty of discriminating among highly confusable candidates, rather than by long-context limitations alone.
Our diagnostics also uncover a recurring failure mode in which models default to positional heuristics when deterministic, content-based selection becomes uncertain.
Taken together, these results motivate evaluation settings that complement broad subject coverage with explicit stress tests for semantic interference and decision reliability.
Such benchmarks better align with deployment conditions, where the effective search space is large and confusable alternatives are the norm.

\section*{Limitations}
Our benchmark is intentionally constructed around a single, high-precision Korean orthography error identification task, chosen to induce dense interference among highly confusable alternatives. 
While this setting is effective for stress-testing reliability under massive-option selection, it remains an open question whether the same failure modes and metric trends hold for other task families such as multi-step reasoning, knowledge-intensive QA, instruction following, or multilingual settings. Establishing the degree of cross-task and cross-domain transfer is therefore an important direction for future work.

Our evaluation design uses a fixed set of targets with repeated Monte Carlo trials over distractor sampling and option sizes $N \in \{4,10,20,50,100\}$. This yields controlled estimates of sensitivity to option density, but it may not capture the full diversity of naturally occurring errors, broader topic coverage, or larger effective candidate spaces encountered in real deployments. In addition, the dataset construction relies on an automatic spell-checker ensemble with strict unanimity filtering and conservative exclusions to prioritize label precision. This choice may systematically remove convention-sensitive or borderline cases that are practically meaningful, potentially biasing the benchmark toward clearer-cut phenomena.

Finally, our model coverage is limited. Due to budget constraints, we evaluated only a subset of frontier models and could not include several other state-of-the-art proprietary and open-weight systems. As a result, our findings should be interpreted as evidence about relative behaviors within the tested model set rather than a comprehensive ranking of frontier capabilities. Broader evaluation across a wider range of frontier LLMs is necessary to strengthen generality and to determine whether the observed behaviors persist across architectures and training regimes.

\bibliography{custom}

\appendix

\section{Spell-Checker Ensemble Labeling}
\label{sec:appendix_ensemble}

\paragraph{Overview and retention.}
We label items using an ensemble of three Korean spell checkers (\textsc{Daum}, \textsc{Saramin}, \textsc{Nara}) and keep only cases with unanimous judgments.
We further exclude convention-sensitive variants described below.
From an initial pool of $1{,}002$ candidates, these filters remove $252$ items, leaving $750$ items for dataset construction.

\paragraph{Unanimity criterion.}
We adopt a strict unanimity rule to prioritize label precision.
A sentence is retained as a \textit{distractor} only when all three spell checkers accept it (i.e., none flags an error).
A perturbed sentence is retained as a \textit{target} only when all three spell checkers flag it as erroneous.
Any item that receives mixed judgments across the tools is discarded.

\paragraph{Filtering of convention-sensitive cases.}
To reduce residual ambiguity even under unanimous tool judgments, we additionally remove items governed by exception-heavy conventions, especially convention-sensitive spacing decisions and acceptable orthographic variants (e.g., cases where both spacing and concatenation are widely regarded as permissible in practice).
We also filter candidates that appear non-standard (e.g., dialectal or archaic forms) to avoid style-guide-dependent judgments.
Table~\ref{tab:dropped_sentence_example} summarizes representative cases that we drop despite unanimous tool outputs.

\paragraph{Error-type coverage.}
The retained targets cover diverse orthographic perturbations (e.g., spacing errors, consonant/vowel substitutions, and final-consonant violations), while the conservative filtering prioritizes precision over coverage for evaluation stability.

\begin{table*}[h]
\centering
\small
\setlength{\tabcolsep}{5pt}
\renewcommand{\arraystretch}{1.15}
\begin{threeparttable}
\begin{tabular}{p{2.6cm} p{5.7cm} p{5.8cm}}
\toprule
\textbf{Rule fragment} & \textbf{English translation} & \textbf{Example (both acceptable)} \\
\midrule

\textbf{Auxiliary verb spacing} \tnote{a}
&
Auxiliary predicates are spaced by default. Concatenation is also permitted in limited constructions, notably when the main predicate is followed by the connective ending \textit{-ㅏ/-ㅓ/-ㅕ} or when an adnominal form is followed by an auxiliary construction such as \textit{-듯하다}, \textit{-만하다}, \textit{-법하다}, \textit{-성싶다}, or \textit{-척하다}. Some forms are obligatorily concatenated, while phrase-level attachments require spacing.
&
\ttfamily\parbox[t]{5cm}{
그는 사과를 \textcolor{red}{먹어 보았다}.\\
그는 사과를 \textcolor{red}{먹어보았다}.}
\\
\addlinespace

\textbf{Permitted concatenation} \tnote{b}
&
Concatenation is permitted in adnominal + auxiliary constructions, yielding widely used paired variants in formal writing.
&
\ttfamily\parbox[t]{5cm}{
비가 \textcolor{red}{올 듯}하다.\\
비가 \textcolor{red}{올듯}하다.}
\\
\addlinespace
\textbf{Standard variants} \tnote{c}
&
When both a full form and its contracted form are widely used and the usefulness of the contracted form is clearly recognized, both are treated as standard.
&
\ttfamily\parbox[t]{5cm}{
저녁\textcolor{red}{노을}이 붉다. / 저녁\textcolor{red}{놀}이 붉다.\\
\textcolor{red}{막대기}로 바닥을 쳤다. / \textcolor{red}{막대}로 바닥을 쳤다.}
\\
\bottomrule
\end{tabular}

\begin{tablenotes}\footnotesize
\item[a] Korean Orthography, Spacing, Article 47. \url{https://www.law.go.kr/LSW/admRulLinkProc.do?admRulNm=%ED%95%9C%EA%B8%80%EB%A7%9E%EC%B6%A4%EB%B2%95&mode=20}
\item[b] NIKL explanations of permitted concatenation patterns for auxiliary predicates. For example, Online Q\&A explanations summarizing the two licensed constructions and the convention-sensitive nature of these pairs. \url{https://www.korean.go.kr/front/onlineQna/onlineQnaView.do?mn_id=261&qna_seq=321299}
\item[c] Standard Language Regulations, Article 16 lists pairs such as \textit{노을/놀} and \textit{막대기/막대}.
\url{https://www.korean.go.kr/front/page/pageView.do?mn_id=94&page_id=P000089}

\end{tablenotes}
\end{threeparttable}
\caption{\textbf{Examples of convention-sensitive forms excluded from spell-checker ensemble labeling due to acceptable orthographic variation.}}
\label{tab:dropped_sentence_example}
\end{table*}

\section{Dataset Difficulty Scoring Heuristic}
\label{sec:appendix_difficulty}

\begin{table*}[t]
\centering
\small
\setlength{\tabcolsep}{6pt}
\renewcommand{\arraystretch}{1.15}
\begin{tabular}{@{}llp{9cm}ll@{}}
\toprule
ID & Feature & Definition / Extraction & Range & Weight $w_k$ \\
\midrule
(i) & $f_{\text{eojeol}}$ & Eojeol count (whitespace-separated tokens). & $N$ & $0.25$ \\
(ii) & $f_{\text{morph}}$ & Morpheme count from a morphological analyzer. & $N$ & $0.25$\\
(iii) & $f_{\text{clause}}$ & \# of clause-marking endings whose surface form is in Table~\ref{tab:clause_endings}. & $N$ & $0.25$\\
(iv) & $f_{\text{conf}}$ & Indicator for high-confusion constructions:
dependent nouns (NNB), auxiliary predicates (VX), sai-siot/compounds. & $\{0,1\}$ & $0.15$\\
(v) & $f_{\text{sym}}$ & Presence of any symbols or numerals. & $\{0,1\}$ & $0.10$\\
\bottomrule
\end{tabular}
\caption{Surface features used in the deterministic difficulty heuristic.}
\label{tab:difficulty_features}
\end{table*}

\begin{table}[h]
\centering
\small
\setlength{\tabcolsep}{10pt}
\renewcommand{\arraystretch}{1.1}
\begin{tabular}{@{}lllll@{}}
\toprule
\multicolumn{5}{c}{Clause-marking endings used for $f_{\text{clause}}$} \\
\midrule
\ttfamily다 & \ttfamily요 & \ttfamily고 & \ttfamily면 & \ttfamily지만  \\
\ttfamily서 & \ttfamily며 & \ttfamily는데 & \ttfamily는다 & \ttfamily습니다 \\
\bottomrule
\end{tabular}
\caption{Example of clause-marking endings (surface forms) counted in $f_{\text{clause}}$.}
\label{tab:clause_endings}
\end{table}

\begin{table*}[h]
\centering
\small
\setlength{\tabcolsep}{5pt}
\renewcommand{\arraystretch}{1.15}
\begin{tabularx}{\textwidth}{X c c c c c c}
\toprule
Sentence & Difficulty Level & Difficulty Score & Eojeol & Morph & Dep.\ Noun & Aux.\ Pred \\
\midrule
\ttfamily\textit{저는 매일 물을 열 컵씩 마십니다.} & 1 & 0.1184 & 6 & 11 & $0$ & $0$ \\
\ttfamily\textit{오늘은 하루 종일 바빴습니다.} & 1 & 0.1217 & 4 & 8  & $0$ & $0$  \\
\midrule
\ttfamily\textit{시간이 없어서 나는 거의 뛰다시피 급히 걸어갔다.} & 2 & $0.2234$ & $7$ & $14$ & $0$ & $0$ \\
\ttfamily\textit{그는 자신이 맡은 일에서만큼은 최고의 전문가다.} & 2 & $0.2239$ & $6$ & $16$ & $1$ & $0$ \\
\midrule
\ttfamily\textit{8시니까 벌써 저녁은 다 먹었다시피 했을 거야.} & 3 & $0.3454$ & $7$ & $18$ & $1$ & $1$ \\
\ttfamily\textit{우리가 팔 걷어붙이고 달려들면 금방 끝낼 일이다.} & 3 & $0.3484$ & $7$ & $14$ & $0$ & $0$ \\
\midrule
\ttfamily\textit{병원 측은 더 이상의 방법이 없다며 나머지는 하늘에 맡기자고 했다.} & 4 & $0.4751$ & $10$ & $20$ & $1$ & $0$ \\
\ttfamily\textit{정밀하게 떨어지는 모래시계의 초침 소리 같은 흙의 속삭임이 귓가에 바짝 가까이 들려온다.} & 4 & $0.4805$ & $12$ & $24$ & $0$ & $1$ \\
\bottomrule
\end{tabularx}
\caption{\textbf{Example sentences by difficulty level under the heuristic scoring scheme}, showing the associated score and selected morpho-syntactic indicators (eojeol/morpheme counts, dependent nouns, and auxiliary predicates).}
\label{tab:difficulty_examples}
\end{table*}

\paragraph{Feature extraction.}
We use the \textsc{Kiwi} morphological analyzer from \texttt{kiwipiepy} (v0.22.1) on each sentence and compute:
(i) eojeol count,
(ii) morpheme count,
(iii) number of clause-marking endings (example list in Table~\ref{tab:clause_endings}),
(iv) indicator for high-confusion constructions (dependent nouns, auxiliary predicates, sai-siot/compounds),
and (v) presence of symbols or numerals.

\paragraph{Scoring.}
For each feature, we apply min--max normalization to $[0,1]$ over the full candidate pool used for dataset construction, and compute
$score(x)=\sum_k w_k \tilde f_k(x)$ with weights $w_k$.
We assign difficulty Levels 1--4 by quartiles of the resulting score distribution.

Table~\ref{tab:difficulty_features} reports the surface features and weights we used in the difficulty evaluation.
Weights are chosen to reflect the relative stability of each signal as a proxy for surface complexity. 
Length- and clause-related features (eojeol count, morpheme count, and clause-marking endings) capture broadly applicable increases in morpho-syntactic load and thus receive the majority of the mass ($0.75$ in total, equally split).
In contrast, the binary flags for high-confusion constructions and symbols/numerals are sparse and can produce abrupt score changes; we therefore downweight them ($0.15$ and $0.10$) so they act as corrective cues rather than primary drivers.

Table~\ref{tab:difficulty_examples} shows representative sentences for each difficulty level, along with their heuristic scores and key surface indicators (eojeol/morpheme counts, dependent nouns, and auxiliary predicates).

\section{Padding and Long-Context Control}
\label{sec:appendix_padding}

\paragraph{Motivation.}
At large option sizes, the serialized prompt becomes long even without explicit padding, so performance drops at high $N$ could be attributed either to (i) genuine massive-option interference (dense, confusable distractors) or (ii) long-context effects from increased prompt length.
To disentangle these factors, we introduce \emph{padding-based length controls} that increase context length while keeping the option set and answer identity unchanged.

\paragraph{Prompt-length equivalence.}
Table~\ref{tab:padding-token-counts} reports average prompt token counts (computed with the \texttt{cl100k\_base} tokenizer) for the no-padding baseline and several padding types.
Across both environments and option sizes, padding expands the context into an approximately 2k--5k token regime, matching or exceeding the unpadded long-context regime induced by large $N$.
Because front/back placement changes only where padding is inserted (not its content), it yields identical token counts; thus any performance differences between front vs.\ back reflect positional/context interactions rather than length differences.

\begin{table}[h]
\centering
\small
\setlength{\tabcolsep}{6pt}
\renewcommand{\arraystretch}{1.15}
\resizebox{\columnwidth}{!}{%
\begin{tabular}{llccccc}
\toprule
\textbf{Env} & \textbf{Condition} & \textbf{N=4} & \textbf{N=10} & \textbf{N=20} & \textbf{N=50} & \textbf{N=100} \\
\midrule
\multirow{5}{*}{Easy} 
& No padding   & 262 & 392 & 611 & 1267 & 2390 \\
& Korean Text padding & 2910 & 3040 & 3259 & 3916 & 5010 \\
& English Text padding & 2244 & 2374 & 2593 & 3250 & 4344 \\
& Symbol padding & 2671 & 2802 & 3018 & 3678 & 4772 \\
& Korean List padding  & 1916 & 2046 & 2256 & 2921 & 4016 \\
\midrule
\multirow{5}{*}{Full} 
& No padding   &  285 & 460 & 749 & 1617 & 3084 \\
& Korean Text padding & 2394 & 3109 & 3398 & 4266 & 5733 \\
& English Text padding & 2268 & 2443 & 2732 & 3600 & 5067 \\
& Symbol padding & 2693 & 2868 & 3157 & 4025 & 5492 \\
& Korean List padding  & 1937 & 2112 & 2404 & 3269 & 4736 \\
\bottomrule
\end{tabular}%
}
\caption{
\textbf{Average prompt token counts} for the no-padding baseline and padding-based length controls at selected option sizes $N$, reported for the Easy and Full environments.
Token counts are reported using the \texttt{cl100k\_base} tokenizer as a consistent length proxy, and front/back placements yield exactly identical counts.
Across $N$, all padding increases context length to approximately 2k--5k tokens, matching or exceeding the unpadded long-context regime induced by large $N$.
}
\label{tab:padding-token-counts}
\end{table}

\begin{table*}[h]
\centering
\small
\setlength{\tabcolsep}{6pt}
\renewcommand{\arraystretch}{1.10}
\resizebox{\textwidth}{!}{%
\begin{tabular}{l l r r r r r}
\toprule
Env & Padding content (min over front/back) 
& Gemini-3-Pro-preview & Gemini-2.5-Flash & A.X-4.0 & HyperCLOVAX-Think & EXAONE-4.0-32B \\
\midrule
\multirow{4}{*}{Easy}
& Korean Text & 0.001 & \textbf{-0.011} & -0.013 & -0.094 & -0.050 \\
& English Text & 0.001 & 0.014 & \textbf{-0.019} & \textbf{-0.482} & \textbf{-0.191} \\
& Random Symbol  & 0.000 & 0.009 & -0.004 & -0.155 & -0.052 \\
& Korean Meaningless List  & 0.000 & 0.009 & 0.001 & -0.019 & -0.068 \\
\midrule
\multirow{4}{*}{Full}
& Korean Text & 0.038 & \textbf{-0.002} & -0.025 & 0.001 & 0.012 \\
& English Text & 0.051 & -0.001 & \textbf{-0.047} & \textbf{-0.172} & \textbf{-0.028} \\
& Random Symbol  & 0.011 & 0.009 & -0.006 & -0.046 & -0.025 \\
& Korean Meaningless List  & 0.025 & -0.001 & -0.029 & 0.046 & 0.012 \\
\bottomrule
\end{tabular}%
}
\caption{\textbf{Which padding content drives the worst-case drop at $N=100$.}
Entries are mean $\Delta=\mathrm{Acc}_{100}^{\text{pad}}-\mathrm{Acc}_{100}^{\text{no-pad}}$.
For each content type, we report the worse of front vs.\ back placement (min over front/back).
Negative values indicate degradation under padding.
Notably, English-prose padding disproportionately harms HyperCLOVAX-Think (and EXAONE on Easy), explaining the large padding variability observed in Figure~\ref{fig:padding_envelope}.}
\label{tab:padding_content}
\end{table*}

Performance robustness under padding.We evaluate multiple padding types, including domain-irrelevant Korean prose, English translated text, symbolic noise, and a meaningless Korean list, alongside two placement strategies: front and back. We provide the exact padding excerpts used for each type in Table~\ref{tab:padding_examples}.

Across models, performance trends remain largely stable across these padding variants. This stability indicates that the major degradations observed at high $N$, particularly in the \textit{Full} environment, are not explained solely by increased prompt length. When deviations occur, they are typically model-specific, such as a sensitivity to English-text padding. Such cases suggest that performance drops stem from content-dependent interactions rather than a generic long-context failure.

\newcolumntype{L}[1]{>{\raggedright\arraybackslash}p{#1}}
\newcolumntype{Y}{>{\raggedright\arraybackslash}X}

\begin{table*}[t]
\centering
\small
\setlength{\tabcolsep}{5pt}
\renewcommand{\arraystretch}{1.12}
\begin{threeparttable}
\begin{tabularx}{\textwidth}{L{3.2cm} L{2.2cm} Y}
\toprule
\textbf{Type} & \textbf{Placement} & \textbf{Excerpt (verbatim, truncated)} \\
\midrule

\vspace{0pt}Korean prose &
\vspace{0pt}Front, Back &
{\ttfamily\footnotesize\vspace{0pt}
천문학, 또는 천체학은 별이나 행성, 혜성, 은하와 같은 천체와, 지구 대기의 바깥쪽으로부터 비롯된 현상을 연구하는 자연과학의 한 분야이다.
우주의 시작 및 진화, 천체의 운동, 물리, 화학, 기상, 진화 등을 연구 대상으로 한다.\newline
역사적으로 천문학은 역법, 천문항법, 점성술까지 수많은 분야들을 포함했다. \newline
천체의 움직임에서 가장 비중이 큰 힘은 중력이므로, 일반상대론을 많이 이용한다.\newline
...
} \\

\addlinespace

\vspace{0pt}English prose\newline
(Translated version of Korean prose) &
\vspace{0pt}Front, Back &
{\ttfamily\footnotesize\vspace{0pt}
Astronomy is a natural science that studies celestial objects and the phenomena that occur in the cosmos. It uses mathematics, physics, and chemistry to explain their origin and their overall evolution. \newline
Objects of interest include planets, moons, stars, nebulae, galaxies, meteoroids, asteroids, and comets. \newline
...
} \\

\addlinespace

\vspace{0pt}Symbolic noise &
\vspace{0pt}Front, Back &
{\ttfamily\footnotesize\vspace{0pt}
\# \# \# \# \# \# \# \# \# \# \# \# \# \newline
\% \% \% \% \% \% \% \% \% \% \% \% \% \newline
\_ \_ \_ \_ \_ \_ \_ \_ \_ \_ \_ \_ \_ \newline
...
} \\

\addlinespace

\vspace{0pt}Lorem Ipsum list (Korean) &
\vspace{0pt}Front, Back &
{\ttfamily\footnotesize\vspace{0pt}
1) 항목 A: 임의 문장입니다.\newline
2) 항목 B: 임의 문장입니다.\newline
...\newline
100) 임의 기록: 내용 없음.
} \\

\bottomrule
\end{tabularx}

\caption{\textbf{Padding files used in the padding-control experiment.} Each file is applied in both front and back positions.}
\label{tab:padding_examples}
\end{threeparttable}
\end{table*}


\section{English-Text Padding Sensitivity of HyperCLOVAX}
\label{sec:appendix_clovax_entext}

We further examine the model-specific exception observed for \textsc{HyperCLOVAX-Think} under padding-based controls.
Across padding conditions, the largest degradation is consistently induced by English-text padding at $N=100$, whereas other padding types yield comparatively smaller shifts (Table~\ref{tab:padding_content}).
This pattern indicates that the robustness failure is not a generic function of prompt length, but reflects a content-dependent vulnerability to specific contextual distributions.

To make this effect concrete, Table~\ref{tab:clovax_entext_mech} contrasts \textsc{HyperCLOVAX-Think} in the \textit{Easy} environment under the no-padding baseline versus English-text padding.
English-text padding reduces $Acc_{100}$ from $0.800$ to $0.649$ and increases the Bubble Index (from $0.190$ to $0.312$), reflecting a larger gap between low-option and massive-option competence under this distribution shift.
Crucially, the accuracy drop is accompanied by systematic changes in positional behavior.
Excess primacy mass rises sharply ($\Delta PFI_{10}$ from $0.091$ to $0.177$), the mean response position shifts toward earlier options ($50.30$ to $44.49$), and the global distributional mismatch increases (KS from $0.111$ to $0.243$).
Together, these diagnostics suggest that English-text padding does not merely add length, but destabilizes candidate ranking and increases reliance on early-option heuristics under uncertainty.

\paragraph{Why the effect is larger in the \textit{Easy} environment.}
The English-padding sensitivity is most pronounced in the \textit{Easy} setting, which is consistent with a combination of headroom effects and a threshold-like policy shift under uncertainty.
In \textit{Easy}, \textsc{HyperCLOVAX-Think} starts from relatively high baseline performance without padding, leaving substantial room for accuracy to drop when the prompt distribution changes.
In \textit{Full}, baseline uncertainty is already higher due to denser and more confusable distractors, so additional degradation from padding can appear comparatively bounded.

More importantly, the \textit{Easy} environment may enable lightweight elimination-style heuristics that remain effective when the context is homogeneous, but become unstable under a strong distribution shift such as English-text padding.
Once this perturbation pushes the model past an uncertainty threshold, it may abandon content-based discrimination and revert to positional fallback behaviors.
This interpretation aligns with the diagnostics in Table~\ref{tab:clovax_entext_mech}, where English padding coincides with a sharp increase in excess primacy mass and a left shift in mean response position, indicating a systematic move toward early-option selection rather than a uniform loss of competence.

\begin{table}[h]
\centering
\small
\setlength{\tabcolsep}{5pt}
\renewcommand{\arraystretch}{1.08}
\begin{tabular}{lcc}
\toprule
\textbf{Metric} & \textbf{No padding} & \textbf{English Text padding (Easy)} \\
\midrule
$Acc_{100}(\uparrow)$ & 0.800 & 0.649 \\
$NA_{100}(\uparrow)$ & 0.798 & 0.646 \\
$BI(\downarrow)$ & 0.190 & 0.312 \\
$\Delta PFI_{10}$ & 0.091 & 0.177 \\
Entropy$_N$ & 0.923 & 0.902 \\
Front$_{20}$ & 0.221 & 0.343 \\
Tail$_{20}$ & 0.195 & 0.196 \\
MeanPos & 50.299 & 44.493 \\
$KS(\downarrow)$ & 0.111 & 0.243 \\
gold\_slope$_{100}$ & -0.014 & -0.012 \\
\bottomrule
\end{tabular}
\caption{\textbf{Mechanism of English-text padding sensitivity for \textsc{HyperCLOVAX-Think} at $N=100$ in the \textit{Easy} environment.}
Relative to the no-padding baseline, English padding coincides with increased excess primacy mass and a left-shift in mean response position, together with a larger KS distance from the empirical gold distribution.}
\label{tab:clovax_entext_mech}
\end{table}

\section{Uncertainty Quantification}
\label{sec:uncertainty}

\paragraph{Sampling variability and target-level confidence intervals.}
We quantify uncertainty in chance-normalized accuracy (NA) from two complementary sources, separating stochasticity induced by the evaluation protocol from heterogeneity across target instances.

First, we estimate \textit{sampling variability} by constructing two-sided 95\% confidence intervals over the $K=1{,}000$ independent trials for each target and option size $N$.
Because each trial redraws distractors and reshuffles the option order, this interval captures sensitivity to distractor composition and target placement under the dynamic protocol. 

Second, we estimate \textit{target heterogeneity} by averaging performance across the $1{,}000$ trials for each target to obtain a stable target-level score, and then forming two-sided 95\% confidence intervals across the $|T|=30$ target sentences.
This view reflects that some orthographic errors are intrinsically easier or harder regardless of the sampled distractors/

Because these procedures weight sources of variation differently, reported mean NA can differ slightly between sampling-based summaries and target-based summaries.
Intervals are not clipped to $[0,1]$ to preserve symmetry and avoid boundary artifacts for near-ceiling models.

\begin{table*}[t]
\centering
\small
\setlength{\tabcolsep}{4.0pt}
\renewcommand{\arraystretch}{1.08}
\resizebox{\textwidth}{!}{%
\begin{tabular}{l l l l l l l}
\toprule
Env & Model & NA$_4$ (95\% CI) & NA$_{10}$ (95\% CI) & NA$_{20}$ (95\% CI) & NA$_{50}$ (95\% CI) & NA$_{100}$ (95\% CI) \\
\midrule
\multirow{5}{*}{Easy}
& Gemini-3-Pro-preview   & \textbf{0.9996} [0.9987, 1.0005] & \textbf{0.9991} [0.9971, 1.0010] & \textbf{0.9979} [0.9953, 1.0005] & \textbf{0.9925} [0.9829, 1.0021] & 0.9898 [0.9713, 1.0083] \\
& Gemini-2.5-Flash       & 0.9937 [0.9840, 1.0034] & 0.9926 [0.9840, 1.0006] & 0.9956 [0.9890, 1.0008] & 0.9897 [0.9753, 1.0041] & 0.9819 [0.9555, 1.0083] \\
& A.X-4.0            & 0.9689 [0.9280, 1.0098] & 0.9573 [0.9002, 1.0144] & 0.9487 [0.8790, 1.0183] & 0.9364 [0.8545, 1.0183] & 0.9508 [0.8841, 1.0176] \\
& HyperCLOVAX-Think      & 0.9853 [0.9677, 1.0030] & 0.9583 [0.9100, 1.0065] & 0.9423 [0.8708, 1.0139] & 0.9057 [0.8029, 1.0085] & 0.7978 [0.6579, 0.9376] \\
& EXAONE-4.0-32B         & 0.9977 [0.9950, 1.0003] & 0.9928 [0.9827, 1.0029] & 0.9747 [0.9399, 1.0096] & 0.9069 [0.8368, 0.9770] & 0.7115 [0.5777, 0.8452] \\
\midrule
\multirow{5}{*}{Full}
& Gemini-3-Pro-preview   & \textbf{0.9842} [0.9610, 1.0073] & \textbf{0.9616} [0.9212, 1.0020] & \textbf{0.9423} [0.8850, 0.9996] & \textbf{0.8879} [0.7996, 0.9762] & \textbf{0.8557} [0.7669, 0.9446] \\
& Gemini-2.5-Flash       & 0.9640 [0.9322, 0.9958] & 0.9479 [0.9072, 0.9886] & 0.9181 [0.8590, 0.9772] & 0.8354 [0.7458, 0.9250] & 0.7144 [0.6035, 0.8252] \\
& A.X-4.0            & 0.8530 [0.7432, 0.9628] & 0.7759 [0.6528, 0.8990] & 0.7309 [0.5948, 0.8670] & 0.6518 [0.4963, 0.8073] & 0.5615 [0.3849, 0.7381] \\
& HyperCLOVAX-Think      & 0.9009 [0.8094, 0.9924] & 0.7922 [0.6623, 0.9221] & 0.6984 [0.5455, 0.8513] & 0.5146 [0.3655, 0.6637] & 0.2043 [0.1100, 0.2985] \\
& EXAONE-4.0-32B         & 0.7492 [0.6309, 0.8675] & 0.6014 [0.4400, 0.7629] & 0.4419 [0.2560, 0.6277] & 0.2806 [0.1361, 0.4250] & 0.1314 [0.0356, 0.2272] \\
\bottomrule
\end{tabular}%
}
\caption{\textbf{Target-level NA across option sizes.} 
Intervals are computed at the \textit{target level} by first averaging NA over the sampled distractor sets for each target sentence and then treating target means as independent samples (intervals are not clipped to $[0,1]$).}
\label{tab:target_level_ci}
\end{table*}

\section{Additional Decision Policy Analyses}
\label{sec:additional-pos-analysis}

We provide additional diagnostics that characterize decision policy under uncertainty at $N=100$.
All positional summaries use an empirical gold-position reference computed from the realized gold indices.
Although the expected gold position is uniform under random permutations, finite sampling can introduce mild deviations from uniformity.
Using the empirical gold reference ensures that deviation measures reflect model behavior beyond these realized permutation fluctuations.

\begin{figure}[t]
  \centering
  \includegraphics[width=\columnwidth]{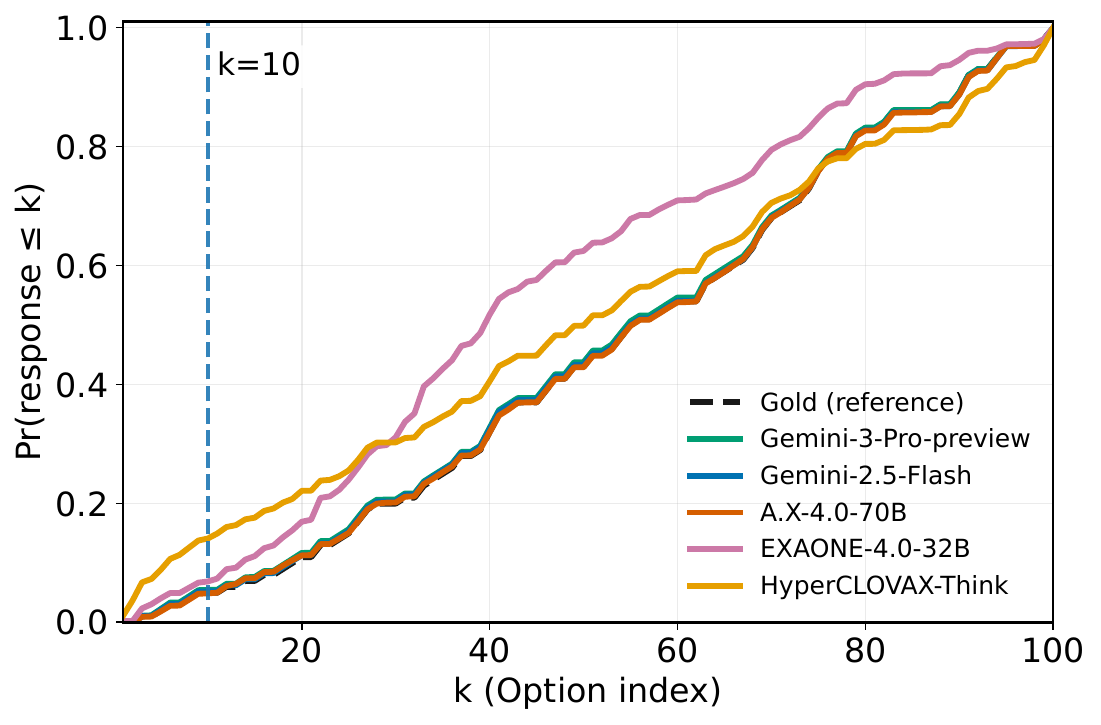}
    \caption{
    \textbf{Response-position densities in \textit{Easy} environment at $N=100$.}
    The gold curve shows the empirical gold-position distribution induced by realized option permutations.
    Deviations from the gold curve indicate positional preference beyond what is explained by finite-sample permutation variability.
    }

  \label{fig:resp_cdf_easy}
\end{figure}

Figure~\ref{fig:resp_cdf_easy} shows that on Easy, response-position distributions remain close to the empirical gold reference, indicating that dense-option formatting alone does not induce systematic position-based collapse when distractors are less confusable.

\begin{figure}[h]
  \centering
  \includegraphics[width=\columnwidth]{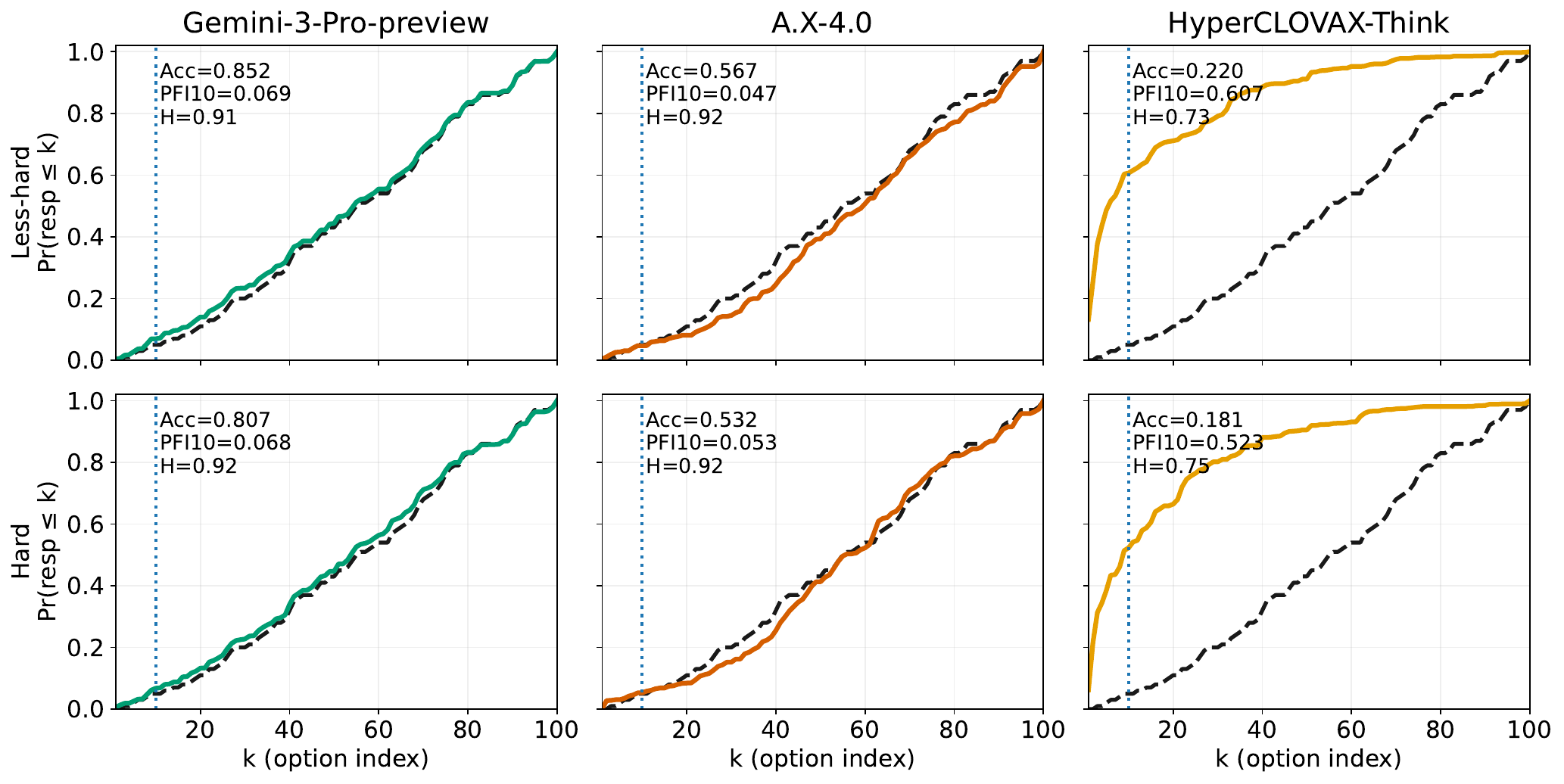}
  \caption{
  \textbf{Decision policy under uncertainty at $N=100$ in the \textit{Full} environment.}
  We split distractor sets into \textit{hard} and \textit{less-hard} groups using trial-level mean accuracy averaged across models.
  Each panel shows the response-position CDF and the empirical gold reference for a representative model.
  }
  \label{fig:policy_case_study}
\end{figure}

Figure~\ref{fig:policy_case_study} conditions on distractor-set difficulty to illustrate that policy differences persist within the same evaluation regime.
High-entropy, near reference-tracking behavior corresponds to broadly distributed selections across indices, while low-entropy prefix concentration corresponds to a fallback policy that allocates most probability mass to early options.

\begin{table*}[t]
\centering
\small
\setlength{\tabcolsep}{5.2pt}
\renewcommand{\arraystretch}{1.08}
\resizebox{\textwidth}{!}{%
\begin{tabular}{l l r r r r r r r r r}
\toprule
Env & Model & $Acc_{100}(\uparrow)$ & $NA_{100}(\uparrow)$ & $\Delta PFI_{10}$ & Entropy$_N(\uparrow)$ & Front$_{20}$ & Tail$_{20}$ & MeanPos & KS$(\downarrow)$ & Slope$_{100}$ \\
\midrule
Easy & Gemini-3-Pro-preview   & \textbf{0.989} & \textbf{0.989} &  0.004 & 0.895 & 0.117 & 0.167 & 54.668 & 0.008 & -0.002  \\
Easy & Gemini-2.5-Flash       & 0.982 & 0.982 &  0.001 & 0.896 & 0.113 & 0.171 & 55.020 & \textbf{0.005} & -0.001  \\
Easy & A.X-4.0            & 0.951 & 0.951 & -0.001 & 0.904 & 0.113 & 0.173 & 55.189 & \textbf{0.005} & 0.008  \\
Easy & HyperCLOVAX-Think      & 0.800 & 0.798 &  0.091 & 0.923 & 0.221 & 0.195 & 50.299 & 0.111 & -0.014  \\
Easy & EXAONE-4.0-32B         & \textcolor{red}{0.714} & \textcolor{red}{0.711} &  0.019 & 0.922 & 0.169 & 0.095 & 45.123 & \textcolor{red}{0.206} & -0.061  \\
\midrule
Full & Gemini-3-Pro-preview   & \textbf{0.827} & \textbf{0.825} &  0.018 & 0.934 & 0.135 & 0.168 & 53.334 & \textbf{0.034} & -0.001  \\
Full & Gemini-2.5-Flash       & 0.717 & 0.714 &  0.025 & 0.950 & 0.151 & 0.166 & 52.305 & 0.050 & -0.008  \\
Full & A.X-4.0            & 0.566 & 0.562 &  0.015 & 0.948 & 0.097 & 0.204 & 56.310 & 0.051 & 0.028  \\
Full & HyperCLOVAX-Think      & 0.212 & 0.204 &  0.526 & 0.770 & 0.705 & 0.018 & 16.936 & \textcolor{red}{0.613} & -0.063  \\
Full & EXAONE-4.0-32B         & \textcolor{red}{0.140} & \textcolor{red}{0.131} &  0.312 & 0.843 & 0.593 & 0.023 & 21.770 & 0.562 & -0.045  \\
\bottomrule
\end{tabular}%
}
\caption{
Positional and policy diagnostics at $N=100$ using the empirical gold-position reference.
$\Delta PFI_{10}$ is excess prefix mass relative to the realized gold positions.
Entropy$_N$ is normalized entropy of the response-position distribution.
KS is the Kolmogorov Smirnov distance between the response and gold CDFs.
Slope$_{100}$ is the count-weighted gold-position slope computed from decile-binned accuracy.
}
\label{tab:positional_appendix}
\end{table*}

\end{document}